\pdfoutput=1
\documentclass[11pt]{article}

\usepackage{authblk}  % added by the author, needed to be imported before anything else

\usepackage{emnlp2021}

\newif\ifanonymous
\anonymousfalse

\usepackage{times}
\usepackage{latexsym}

\usepackage[T1]{fontenc}
\usepackage[utf8]{inputenc}

\usepackage{microtype}

\usepackage{tabularx}
\usepackage{subcaption}
\usepackage{booktabs}
\usepackage[flushleft]{threeparttable}
\usepackage{multirow}
\usepackage{amsmath}
\usepackage{amssymb}
\usepackage[inline]{enumitem}
\setlist{nosep}  % remove spacing between items
\setlist[description]{leftmargin=2ex}  % less hanging indentation for description
\setlist[enumerate]{labelindent=0ex,labelwidth=1.5ex,leftmargin=!}
\setlist[itemize]{labelindent=0ex,labelwidth=1.5ex,leftmargin=!}
\usepackage{colortbl}
\usepackage{mathtools}
\DeclareMathOperator{\argmax}{argmax}
\usepackage[frozencache=true,cachedir=minted-cache]{minted}
\usepackage{xspace}
\usepackage[export]{adjustbox}
\usepackage{xurl}

\usepackage{cleveref}
\crefname{section}{Section}{Sections}
\Crefname{section}{Section}{Sections}
\crefname{appendix}{Appendix}{Appendices}
\crefname{Appendix}{Appendix}{Appendices}
\crefname{subsection}{Section}{Sections}
\Crefname{subsection}{Section}{Sections}
\crefname{equation}{Equation}{Equations}
\Crefname{equation}{Equation}{Equations}
\Crefname{figure}{Figure}{Figures}
\crefname{figure}{Figure}{Figures}
\Crefname{table}{Table}{Tables}
\crefname{table}{Table}{Tables}
\crefname{enumi}{}{}
\usepackage{CJKutf8}
\title{Capturing Logical Structure of Visually Structured Documents with Multimodal Transition Parser}

\author[12]{Yuta Koreeda}
\author[2]{Christopher D. Manning}
\affil[1]{Hitachi America, Ltd, Santa Clara, CA, USA}
\affil[2]{Stanford University, Stanford, CA, USA}
\affil[ ]{{\tt \{koreeda, manning\}@stanford.edu}}
\date{}

\def\textsubsuperscript#1#2{\rlap{\textsuperscript{#1}}\textsubscript{#2}\xspace}
\newcommand{\contractpdfen}{Contract\,\textsubsuperscript{\textit{pdf}}{\textit{en}}\xspace}
\newcommand{\contracttxten}{Contract\,\textsubsuperscript{\textit{txt}}{\textit{en}}\xspace}
\newcommand{\contractpdfja}{Contract\,\textsubsuperscript{\textit{pdf}}{\textit{ja}}\xspace}
\newcommand{\lawpdfen}{Law\,\textsubsuperscript{\textit{pdf}}{\textit{en}}\xspace}
\begin{document}

\maketitle
\begin{abstract}
While many NLP pipelines assume raw, clean texts, many texts we encounter in the wild, including a vast majority of legal documents, are not so clean, with many of them being visually structured documents (VSDs) such as PDFs.
Conventional preprocessing tools for VSDs mainly focused on word segmentation and coarse layout analysis, whereas fine-grained logical structure analysis (such as identifying paragraph boundaries and their hierarchies) of VSDs is underexplored.
To that end, we proposed to formulate the task as prediction of \emph{transition labels} between text fragments that maps the fragments to a tree, and developed a feature-based machine learning system that fuses visual, textual and semantic cues.
Our system is easily customizable to different types of VSDs and it significantly outperformed baselines in identifying different structures in VSDs.
For example, our system obtained a paragraph boundary detection F1 score of 0.953 which is significantly better than a popular PDF-to-text tool with an F1 score of 0.739.

\end{abstract}

\section{Introduction}\label{sec:introduction}

Despite recent motivation to utilize NLP for wider range of real world applications, most NLP papers, tasks and pipelines assume raw, clean texts.
However, many texts we encounter in the wild, including a vast majority of legal documents (e.g., contracts and legal codes), are not so clean, with many of them being visually structured documents (VSDs) such as PDFs.
For example, of 7.3 million text documents found in Panama Papers (which arguably approximates the distribution of data one would see in the wild), approximately 30\% were PDFs\footnote{Calculated from \citet{obermaier_about_2016} by regarding their \textit{emails}, \textit{PDFs} and \textit{text documents} as the denominator.}.
Good preprocessing of VSDs is crucial in order to apply recent advances in NLP to real world applications.

\begin{figure*}
    \begin{center}
        \includegraphics[width=\textwidth]{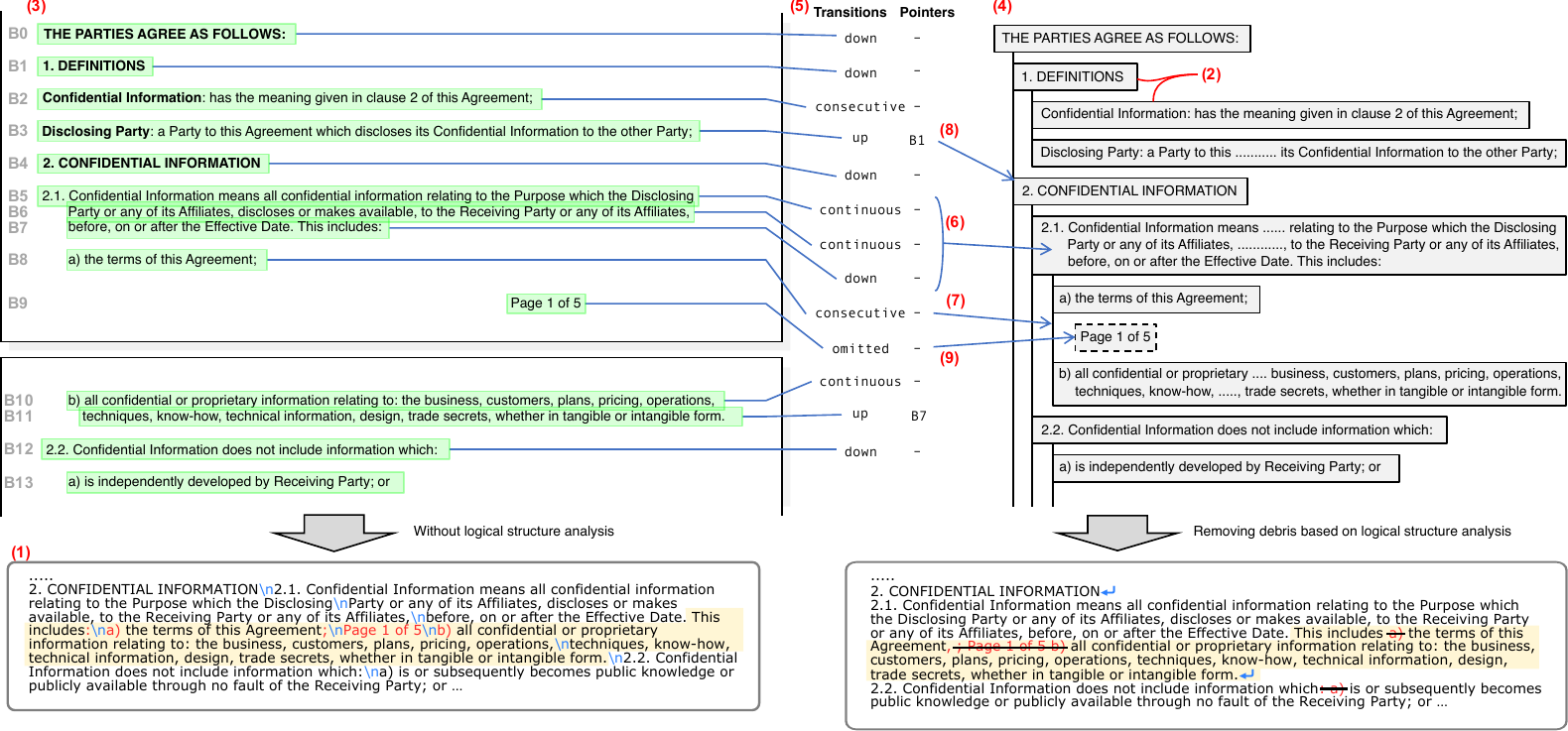}
        \caption{Overview of the logical structure analysis for VSDs and its formulation.}
        \label{fig:overview}
    \end{center}
\end{figure*}

Thus far, the most micro and macro extremes of VSD preprocessing have been extensively studied, such as word segmentation and layout analysis \citep[detecting figures, body texts, etc.;][]{soto_visual_2019,stahl_deeppdf_2018}, respectively.
While these two lines of studies allow extracting a sequence of words in the body of a document, neither of them accounts for local, logical structures such as paragraph boundaries and their hierarchies.

These structures convey important information in any domain, but they are particulary important in the legal domain.
For example, \cref{fig:overview}(1) shows raw text extracted from a non-disclosure agreement (NDA) in PDF format.
An information extraction (IE) system must be aware of the hierarchical structure to successfully identify target information (e.g., extracting ``definition of confidential information'' requires understanding of hierarchy as in \cref{fig:overview}(2)).
Furthermore, we must utilize the logical structures to remove debris that has slipped through layout analysis (``Page 1 of 5'' in this case) and other structural artifacts (such as semicolons and section numbers) for a generic NLP pipeline to work properly.

Yet, such logical structure analysis is difficult.
Even the best PDF-to-text tool with a word-related error rate as low as 1.0\% suffers from 17.0\% newline detection error \cite{bast_benchmark_2017} that is arguably the easiest form of logical structure analysis.

The goal of this study is to develop a fine-grained logical structure analysis system for VSDs.
We propose a transition parser-like formulation of logical structure analysis, where we predict a \emph{transition label} between each consecutive pair of text fragments (e.g., two fragments are in a same paragraph, or in different paragraphs of different hierarchies).
Based on such formulation, we developed a feature-based machine learning system that fuses multimodal cues: visual (such as indentation and line spacing), textual (such as section numbering and punctuation), and semantic (such as language model coherence) cues.
Finally, we show that our system is easily customizable to different types of VSDs and that it significantly outperforms baselines in identifying different structures in VSDs.
For example, our system obtained a paragraph boundary detection F1 score of 0.953 that is significantly better than PDFMiner\footnote{\url{https://euske.github.io/pdfminer/}}, a popular PDF-to-text tool, with an F1 score of 0.739.
We open-sourced our system and dataset\footnote{\url{https://github.com/stanfordnlp/pdf-struct}}.

\section{Problem Setting and Our Formulation}\label{sec:problem_definition}

In this study, we concentrate on logical structure analysis of VSDs.
The input is a sequence of text blocks (\cref{fig:overview}(3)) that can be obtained by utilizing existing coarse layout analysis and word-level preprocessing tools.
We aim to extract paragraphs and identify their relationships.
This is equivalent to creating a tree with each block as a node (\cref{fig:overview}(4)).

We propose to formulate this tree generation problem as identification of a \emph{transition label} between each consecutive pair of blocks (\cref{fig:overview}(5)) that defines their relationship in the tree.
We define the transition $\mathit{trans}_{i}$ between $i$-th block (hereafter $b_i$) and $b_{i+1}$ as one of the following:
\begin{description}[font=\normalfont\ttfamily]
    \item[continuous] \quad $b_i$ and $b_{i+1}$ are continuous in a single paragraph (\cref{fig:overview}(6)).
    \item[consecutive] \quad $b_{i+1}$ is the start of a new paragraph at the same level as $b_i$  (\cref{fig:overview}(7)).
    \item[down] \quad $b_{i+1}$ is the start of a new paragraph that is a child (a lower level) of the paragraph that $b_i$ belongs to  (\cref{fig:overview}(6)).
    \item[up] \quad $b_{i+1}$ is the start of a new paragraph that is in a higher level than the paragraph that $b_i$ belongs to  (\cref{fig:overview}(8)).
    \item[omitted] \quad $i$-th block is debris and omitted (\cref{fig:overview}(9)). $\mathit{trans}_{i-1}$ is carried over to the relationship betwen $b_{i-1}$ and $b_{i + 1}$.
\end{description}

While \texttt{down} is well-defined (because we assume a tree), \texttt{up} can be ambiguous as to how many levels we should raise.
To that end, we also introduce a \textit{pointer} to each \texttt{up} block, which points at $b_{j}$ whose level $b_{i}$ belongs to ($\mathit{ptr}_{i} = b_j$, where $j < i$; \cref{fig:overview}(8)).

\section{Dataset}\label{sec:dataset}

\begin{figure*}[tb]
    \centering
    \begin{subfigure}[b]{0.23\textwidth}
        \centering
        \includegraphics[width=\textwidth,frame]{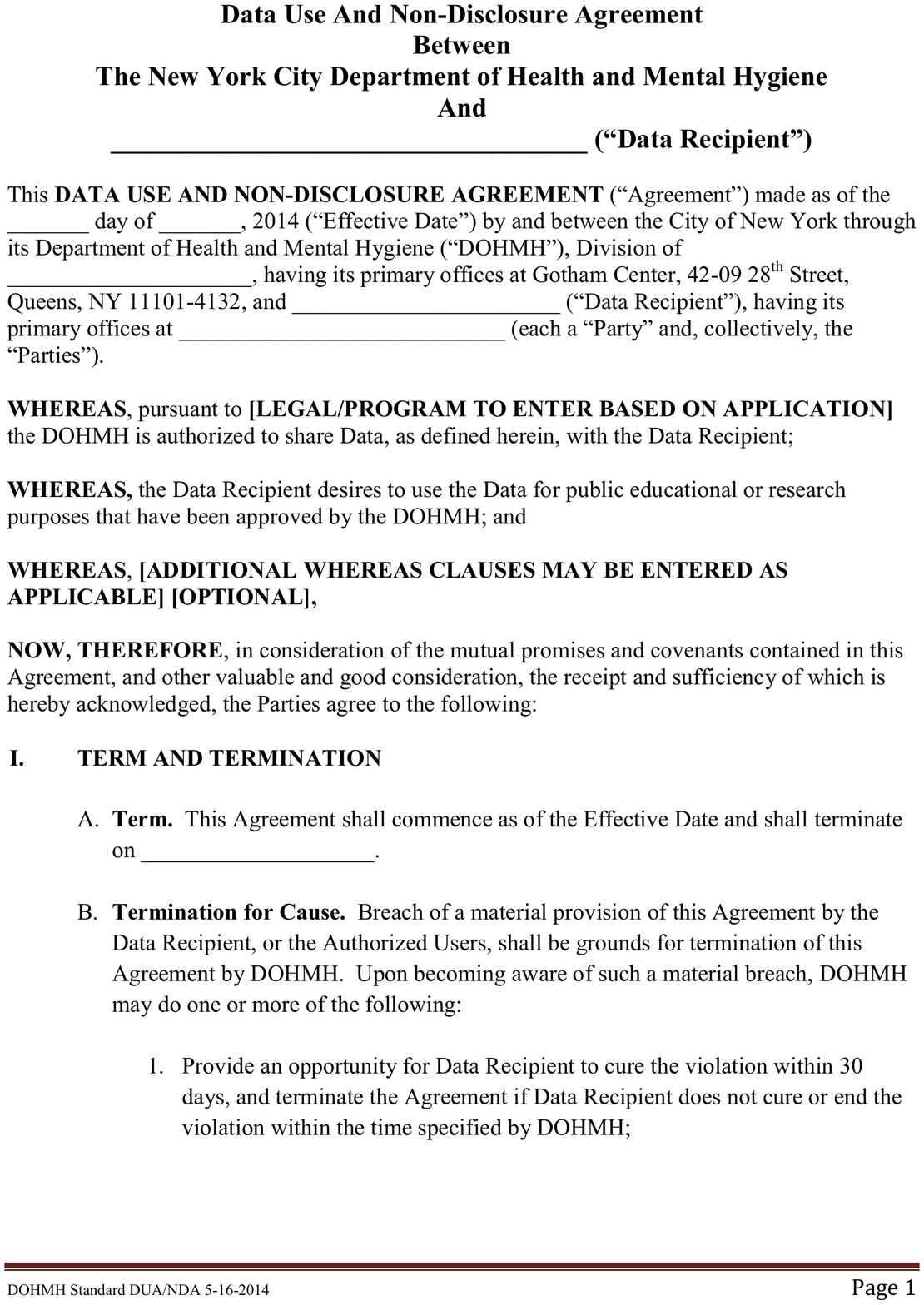}
        \caption{\contractpdfen}
    \end{subfigure}~
    \begin{subfigure}[b]{0.23\textwidth}
        \centering
        \includegraphics[width=\textwidth,frame]{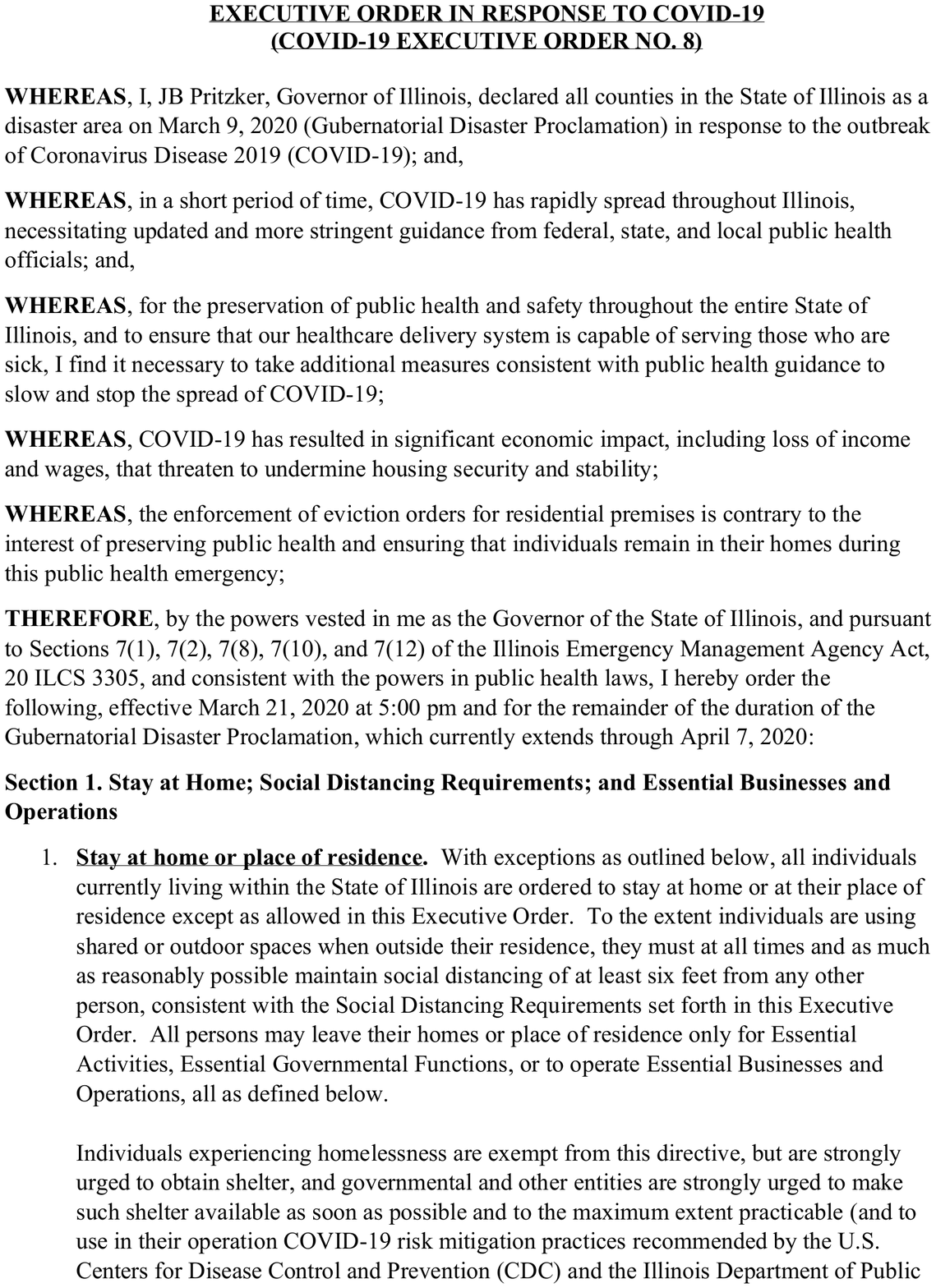}
        \caption{\lawpdfen}
    \end{subfigure}~
    \begin{subfigure}[b]{0.23\textwidth}
        \centering
        \includegraphics[width=\textwidth,frame]{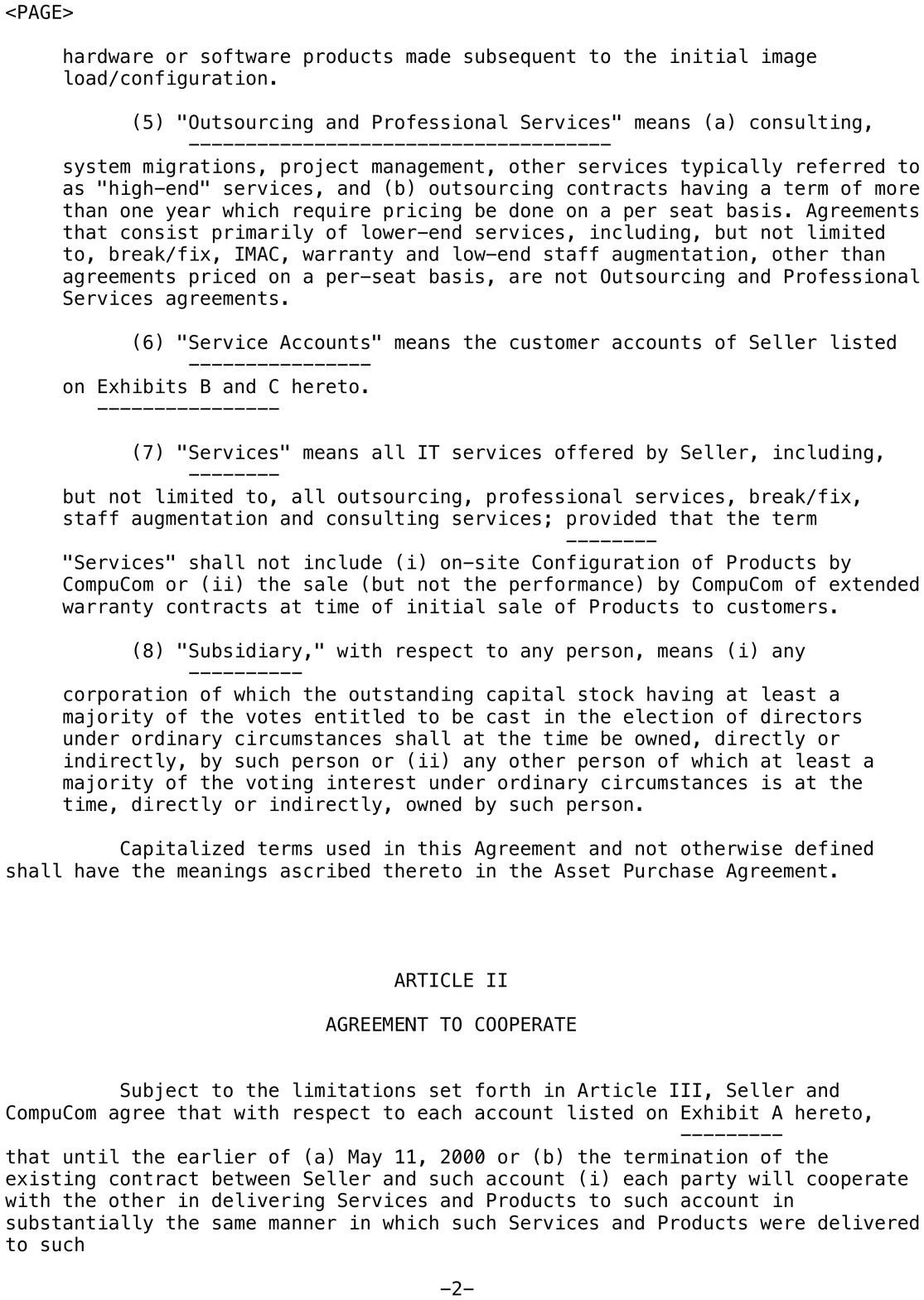}
        \caption{\contracttxten}
    \end{subfigure}~
    \begin{subfigure}[b]{0.23\textwidth}
        \centering
        \includegraphics[width=\textwidth,frame]{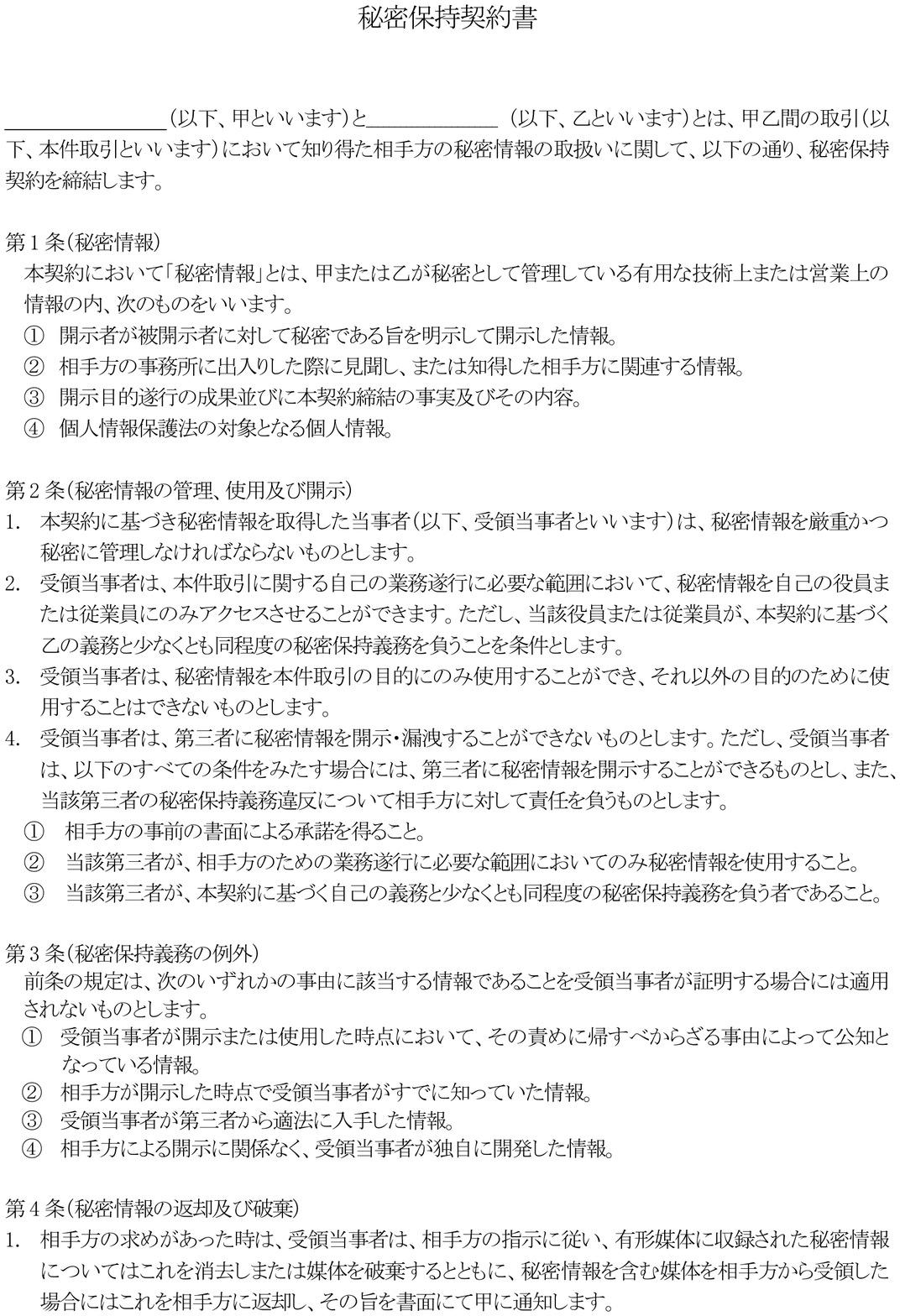}
        \caption{\contractpdfja}
    \end{subfigure}
    \caption{Examples of VSDs in our dataset\protect\footnotemark}\label{fig:example}
\end{figure*}

In this study, we target four types of VSDs in different file formats and languages:
\begin{description}
    \item[\contractpdfen] English NDAs in PDF format.
    \item[\lawpdfen] English executive orders from local authorities.
    \item[\contracttxten] English NDAs in visually structured plain text format.
    \item[\contractpdfja] Japanese NDAs in PDF format.
\end{description}
Examples of each type of VSDs are shown in \cref{fig:example}.

For PDFs, we downloaded PDFs from Google.com search result.
Since our focus is not on coarse layout analysis or word-level preprocessing, we selected single column documents and extracted blocks with an existing software.
\footnotetext{(a) \url{http://www.astho.org/Programs/Infectious-Disease/Healthcare-Associated-Infections/Electronic-Health-Records/Toolkit/Data-Use-Agreement-New-York-City/}, (b) \url{https://www2.illinois.gov/IISNews/21288-Gov._Pritzker_Stay_at_Home_Order.pdf}, (c) \url{https://www.sec.gov/Archives/edgar/data/86115/0000930661/0000930661-99-001321-index.htm}, and (d) \url{http://www.septima.co.jp/contracts/27_himitsuhoji.pdf}}
Specifically, we utilized PDFMiner and extracted each \texttt{LTTextLine}, which roughly corresponds to each line of text, as a block. We merged multiple \texttt{LTTextLine}s where \texttt{LTTextLine}s are vertically overlapping.

For plain texts, we searched documents filed at EDGAR\footnote{\url{https://www.sec.gov/edgar.shtml}}.
We simply used each non-blank line of a plain text as a block.

We annotated all documents by hand.
We describe more details of the data collection and annotation in \cref{sec:appendix-dataset}.

The data statistics are given in \cref{tab:data-statistics}.
While the number of documents is somewhat limited, we note that each document comes with many text blocks and evaluations were stable.
Furthermore, it was enough to reliably show the difference between our system and baselines in our experiments.

\section{Proposed System}\label{sec:method}

\subsection{Transition Parser}\label{sec:method-transition}

In this work, we propose to employ handcrafted features and a machine learning-based classifier as the transition parser.
This strategy is more suited to our task than utilizing deep learning because
\begin{enumerate*}[label=(\arabic*)]
    \item we can incorporate visual, textual and semantic cues, and
    \item it only requires a small number of training data which is critical in the legal domain where most data is proprietary
\end{enumerate*}.

\begin{table}[tb]
    \centering
    \begin{threeparttable}
        \centering
        \fontsize{8pt}{10pt}\selectfont
        \setlength{\tabcolsep}{1pt}
        \renewcommand{\arraystretch}{.8}
        \begin{tabular}{lr@{\hskip 2 pt}lr@{\hskip 2 pt}lr@{\hskip 2 pt}lr@{\hskip 2 pt}l}\toprule
             & \multicolumn{2}{c}{\contractpdfen} & \multicolumn{2}{c}{\lawpdfen} & \multicolumn{2}{c}{\contracttxten} & \multicolumn{2}{c}{\contractpdfja} \\\midrule
             Format & \multicolumn{2}{c}{PDF} & \multicolumn{2}{c}{PDF} & \multicolumn{2}{c}{Text} & \multicolumn{2}{c}{PDF} \\
             Language & \multicolumn{2}{c}{English} & \multicolumn{2}{c}{English} & \multicolumn{2}{c}{English} & \multicolumn{2}{c}{Japanese} \\
             \#Documents & \multicolumn{2}{c}{40} & \multicolumn{2}{c}{40} & \multicolumn{2}{c}{22} & \multicolumn{2}{c}{40} \\
             \midrule
             \#Text blocks & 137.9 & & 165.9 & & 142.0 & & 73.7 &\\
             Max. depth & 3.4 &  & 3.9 &  & 3.1 &  & 3.0 &  \\
             \#\texttt{continuous} & 95.4 & (68\%) & 110.6 & (67\%) & 109.9 & (77\%) & 33.9 & (44\%) \\
             \#\texttt{consecutive} & 20.3 & (17\%) & 30.8 & (20\%) & 15.3 & (12\%) & 14.9 & (20\%) \\
             \#\texttt{up} & 8.5 & (\enspace 6\%) & 7.1 & (\enspace 4\%) & 4.8 & (\enspace 3\%) & 11.0 & (15\%) \\
             \#\texttt{down} & 9.4 & (\enspace 6\%) & 9.9 & (\enspace 6\%) & 4.6 & (\enspace 3\%) & 12.1 & (17\%) \\
             \#\texttt{omitted} & 4.4 & (\enspace 3\%) & 7.6 & (\enspace 3\%) & 7.4 & (\enspace 4\%) & 1.8 & (\enspace 2\%) \\\bottomrule
        \end{tabular}
        \begin{tablenotes}[para,flushleft]
            \raggedright
            \fontsize{8pt}{8pt}\selectfont
           A number in the second set of rows indicates an average count over documents.  A percentage represents an average ratio of each label.
        \end{tablenotes}
    \end{threeparttable}
    \caption{Dataset information}\label{tab:data-statistics}
\end{table}

\begin{table*}[t]
    \centering
    \begin{threeparttable}
        \centering
        \fontsize{8pt}{10pt}\selectfont
        \renewcommand{\arraystretch}{.8}
        \begin{tabular}{p{0pt}>{\raggedleft\arraybackslash}p{2ex}lp{1cm}ccc}\toprule
            &  &  &  & \multicolumn{3}{c}{Document type} \\\cmidrule(lr){5-7}
            &  &  &  & \contractpdfen & &  \\
            &  & Description & Blocks & /\lawpdfen & \contracttxten & \contractpdfja \\\midrule
            \multicolumn{6}{l}{\textbf{V}isual features} \\
            & V1 & Indentation (up, down or same) & 1-2, 2-3 & \checkmark & \checkmark & \checkmark \\
            & V2 & Indentation after erasing numbering & 1-2, 2-3 &  & \checkmark & \\
            & V3 & Centered & 2, 3 & \checkmark & \checkmark & \checkmark \\
            & V4 & Line break before right margin\textsuperscript{\textasteriskcentered} & 1, 2 & \checkmark & \checkmark & \checkmark \\
            & V5 & Page change & 1-2, 2-3 & \checkmark &  & \checkmark\\
            & V6 & Within top 15\% of a page & 2 & \checkmark &  & \checkmark\\
            & V7 & Within bottom 15\% of a page & 2 & \checkmark &  & \checkmark\\
            & V8 & Larger line spacing\textsuperscript{\textasteriskcentered} & 1-2, 2-3  & \checkmark & \checkmark & \checkmark\\
            & V9 & Justified with spaces in middle  & 2, 3 & \checkmark & \checkmark & \checkmark \\
            & V10 & Similar text in a similar position\textsuperscript{\textasteriskcentered} & 2 & \checkmark &  & \checkmark \\
            & V11 & Emphasis by spaces between characters & 1, 2 & &  & \checkmark \\
            & V12 & Emphasis by parentheses & 1, 2 & &  & \checkmark \\[2pt]
            \multicolumn{6}{l}{\textbf{T}extual features} \\
            & T1 & Numbering transition\textsuperscript{\textasteriskcentered} & 2 & \checkmark & \checkmark & \checkmark \\
            & T2 & Punctuated & 1, 2 & \checkmark & \checkmark  & \checkmark \\
            & T3 & List start (\texttt{/[-;:,]\$/}) & 1, 2 & \checkmark & \checkmark & \checkmark \\
            & T4 & List elements (\texttt{/(;|,|and|or)\$/}) & 2 & \checkmark & \checkmark & \\
            & T5 & Page number (strict) & 1, 2, 3 & \checkmark & \checkmark & \checkmark \\
            & T6 & Page number (tolerant) & 1, 2, 3 & \checkmark & \checkmark & \checkmark \\
            & T7 & Starts with ``whereas'' & 3 & \checkmark & \checkmark & \\
            & T8 & Starts with ``now, therefore''  & 3 & \checkmark & \checkmark & \\
            & T9 & Dictionary-like (includes ``:'' \& not V4) & 2, 3& \checkmark & \checkmark & \\
            & T10 & All capital & 2, 3 & \checkmark & \checkmark & \\
            & T11 & Contiguous blank field (underbars) & 1-2, 2-3 & \checkmark & \checkmark & \checkmark \\
            & T12 & Horizontal line (``*-=\#\%\_+'' only) & 1, 2, 3 & & \checkmark & \\[2pt]
            \multicolumn{6}{l}{\textbf{S}emantic features} \\
            & S1 & Language model coherence\textsuperscript{\textasteriskcentered} & 1-2-3 & \checkmark & \checkmark  & \checkmark\\
            \bottomrule
        \end{tabular}
        \makeatletter\def\TPT@hsize{}\makeatletter
        \begin{tablenotes}[para,flushleft]
            \raggedright
            \fontsize{8pt}{8pt}\selectfont
            The ``Blocks'' columns list blocks used to extract features for $\mathit{trans}_{2}$ (e.g. ``1-2, 2-3'' means $\left[b_{i-1}, b_{i}\right]$ and $\left[b_{i}, b_{i+1}\right]$ are used to extract two sets of features). Features with a similar intended functionality are assigned the same feature name and implementations may vary for different document types.
            \textsuperscript{\textasteriskcentered}:~Explained in detail in \cref{sec:method-transition}.
        \end{tablenotes}
    \end{threeparttable}
    \caption{List of features for each feature extractor}\label{tab:features}
\end{table*}

For each block, our parser extracts features from a context of four blocks and performs multi-class classification over the five transition labels.
Since \texttt{omitted} changes targets of transition, we also omit \texttt{omitted} blocks in feature extraction.
For $\mathit{trans}_{i} \neq \texttt{omitted} $, we extract features from $\left[b_{i-1}, b_{i}, b_{j}, b_{j+1}\right]$ where $b_{j}$ is the first block after $b_{i}$ with $\mathit{trans}_{j} \neq \texttt{omitted}$.
For $\mathit{trans}_{i} = \texttt{omitted}$, we extract features from $\left[b_{i-1}, b_{i}, b_{i+1}, b_{i+2}\right]$.
At test time, since we need to know the presence of $\texttt{omitted}$ before feature extraction, we run a first pass of predictions to identify blocks with $\texttt{omitted}$, then use that information to dynamically extract features to identify other labels.

Our system can be customized to different types of documents by modifying the features.
We have designed a feature set for each document type by visually inspecting the training dataset (\cref{tab:features}).
For \contracttxten, we regarded space characters as horizontal spacing and blank lines as vertical spacing, which allowed us to define features that are analogous to those for PDFs.

\begin{figure}
    \begin{center}
        \includegraphics{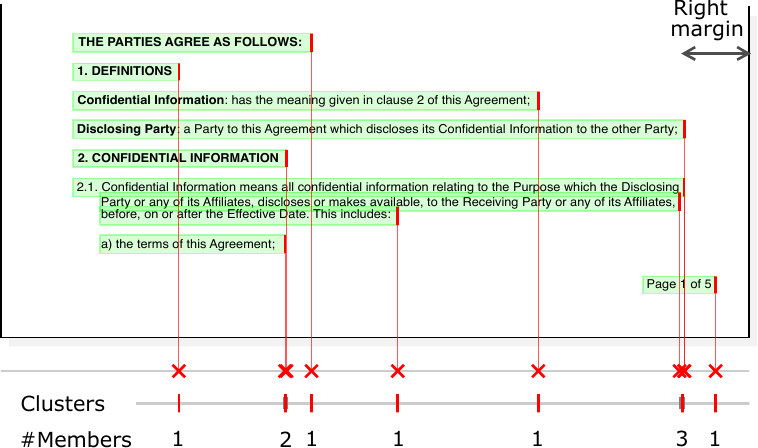}
        \caption{A sketch of how we determine right margin. We apply 1D clustering on the right positions of the blocks and choose the rightmost cluster with at least a user-defined number of members. If we choose to have a minimum of two members, the right margin would be the cluster with three members.}
        \label{fig:margin}
    \end{center}
\end{figure}

While readers can reference our open-sourced code for the concrete implementation, we will discuss some of the features that have important implementation details.
For a target block $b_{i}$:
\begin{description}
        \item[Numbering transition (T1)] A categorical feature that itself is a heuristic transition parser. It identifies a numbering in each block and keeps a memory of the largest numberings by their types (i.e., its alphanumeric type and styling, such as \texttt{IV.} and \texttt{(a)}). It outputs
    \begin{enumerate*}[label=(\arabic*)]
        \item \texttt{continuous} if no numbering is found,
        \item \texttt{consecutive} if the numbering in $b_{i+1}$ is contiguous to the numbering in $b_{i}$,
        \item \texttt{up} if not consecutive and there is a corresponding number in the memory, and
        \item \texttt{down} if it is none of above and it is the first number in its numbering type
    \end{enumerate*}.
    For example, B0 in \cref{fig:overview} is \texttt{down} as \texttt{1.} is the first numbering type that it sees and ``1'' will be added to the memory. B1 and B2 are \texttt{continuous} as no numbering is found and B3 is \texttt{consecutive} as a number ``2'' is found in the same type as \texttt{1.}. B4 is \texttt{down} as it contains a new numbering type.
    \item[Language model coherence (S1)] To determine if $b_{i}$ should be classified as \texttt{omitted}, it utilizes language model to classify whether it is more natural to have $b_{i}$ or $b_{i+1}$ after $b_{i - 1}$. Specifically, we use GPT-2 \cite{radford2019language} to calculate language model loss $\ell(i, i - 1)$ for $b_{i}$ given $b_{i - 1}$ as a context (i.e., fed into the model but not used in the loss calculation). We then calculate $\ell(i, i - 1) - \ell(i + 1, i - 1)$ as the feature. If it is more coherent to have $b_{i}$ after $b_{i - 1}$, $\ell(i, i - 1)$ will be smaller than $\ell(i + 1, i - 1)$ and the feature value will be negative. We also utilize $\ell(i + 1, i) - \ell(i + 1, i - 1)$.
    \item[Similar text in similar position (V10)] Headers and footers tend to appear in similar positions across different pages with similar texts. For example, a contract may have the contract's title on every pages at the same position. This feature is $1$ if there exists a block $b_{j}$ such that blocks' overlapping area is larger than 50\% of their bounding box (treating as if they are on the same page), and their edit distance is small\footnote{$d(b_{i}, b_{j}) / \max(len(b_{i}), len(b_{j})) < 0.1$, where $d$ gives the Levenshtein distance and $len$ gives the length of text.}.
    \item[Line break before right margin (V4)] A~Boolean feature that is $0$ if the block spans to the right margin and $1$ otherwise (i.e., breaks before the right margin). To distinguish the body and the margin of the document, we apply 1D clustering\footnote{We utilized a na\"{\i}ve 1D clustering, where it greedily adds elements from a sorted list to a cluster while the maximum difference of the elements is within a user-defined threshold.} on the right positions of the blocks and extract the rightmost cluster with minimum members of six per page (to ignore headers/footers) as the right margin (\cref{fig:margin}). This margin information is used in other features (V3, V6 and V7).
    \item[Larger line spacing (V8)] A Boolean feature that is $0$ if line spacing is \emph{normal} and $1$ otherwise. To determine the \emph{normal} line spacing, we apply 1D clustering on line spacings and pick a cluster with the largest number of members.
\end{description}

\subsection{Pointer Identification}

We also implement the pointer identification with handcrafted features and a machine learning-based classifier.
Since a \texttt{down} transition creates a new level that a block can point back to, we extract all pairs of $\left[b_{j}, b_{i}\right]$ ($b_j \in \mathcal{C}_{i}$) with $\mathit{trans}_{i} = \texttt{up}$, $\mathit{trans}_{j} = \texttt{down}$ and $j < i$.
We then extract features from $\left[b_{j}, b_{i}\right]$ and train a binary classifier to predict $p\left(\mathit{ptr}_{i} = b_j\middle|b_{j}, b_{i}\right)$.
In training, we use ground truth \texttt{down} labels to extract candidates $\mathcal{C}_{i}$.
At test time, we aggregate $\mathcal{C}_{i}$ from predicted transition labels and predict the pointer by $\mathit{ptr}_{i} = \argmax_{b_j\in\mathcal{C}_{i}} p\left(\mathit{ptr}_{i} = b_j\middle|b_{j}, b_{i}\right)$.

While our pointer points at a block with \texttt{down} ($b_{j}$), it is sometimes important to extract features from the first block in the paragraph that $b_{j}$ belongs to, which we will hereafter refer as $b_{head(j)}$.
Using $b_{head(j)}$, we extract the following features from $\left[b_{j}, b_{i}\right]$:
\begin{description}
    \item[Consecutive numbering] Boolean features on whether a numbering in $b_{i}$ is contiguous to a numbering in $b_{j}$ and $b_{head(j)}$, respectively.
    \item[Indentation] Categorical features on whether indentation gets larger, smaller or stays the same from $b_{j}$ to $b_{i}$ and from $b_{head(j)}$ to $b_{i+1}$, respectively.
    \item[Left aligned] Binary features on whether $b_{j}$, $b_{i+1}$ and $b_{head(j)}$ are left aligned, respectively.
    \item[Transition counts] We count numbers of blocks $\{b_{k}\}_{j < k < i}$ with \texttt{down} and with \texttt{up}, respectively. We use these two numbers along with their difference as features. This is based on an intuition that a closer block with \texttt{down} tends to be more important.
\end{description}
Pointer features are also customizable, but we used the same features\footnote{More precisely, the pointer features are implemented slightly different for different document types, such as numbering being modified to Japanese for \contractpdfja, but they are intended to have similar functionalities.} for all the document types.

\begin{figure}
    \begin{minted}[
        frame=lines,
        framesep=2mm,
        baselinestretch=0.8,
        fontsize=\scriptsize
        ]{python}
class PDFFeatureExtractor(BaseFeatureExtractor):
    def __init__(self, text_boxes):
        bboxes = np.array(
            [tb.bbox for tb in text_boxes])
        page_top = bboxes[:, 3].max()
        page_bottom = bboxes[:, 1].min()
        self.header_thresh = \
            page_top - 0.15 * (page_top - page_bottom)
        ...

    @single_input_feature([1])
    def header_region(self, tb):
        return tb.bbox[3] > self.header_thresh

    @pairwise_feature([(0, 1), (1, 2)])
    def page_change(self, tb1, tb2):
        if tb1 is None or tb2 is None:
            return True
        return tb1.page != tb2.page

    @pointer_feature()
    def pointer_left_aligned(
            self, head_tb, tb1, tb2, tb3):
        return self.left_aligned(tb1)

    ...
    \end{minted}
    \vspace{-2ex}  % ugly workaround
    \caption{The Python implmenetation of a feature extractor}\label{fig:implementation}
\end{figure}

\begin{figure}
    \begin{center}
        \includegraphics{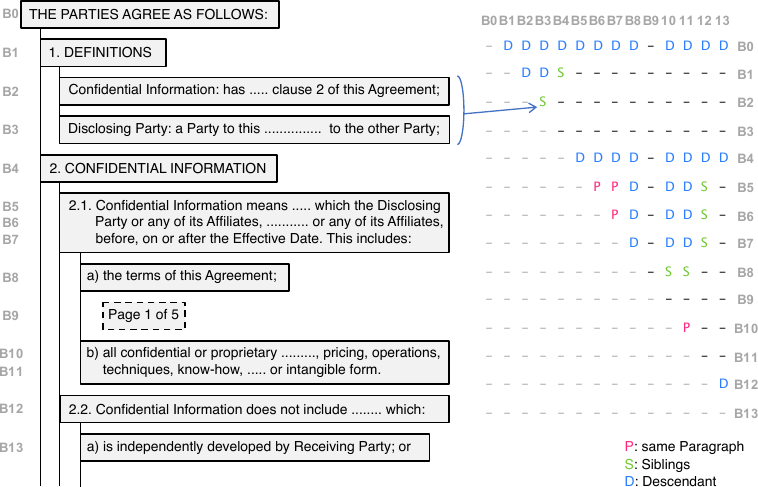}
        \caption{Evaluation from IE perspective. For each of ground truth and predicted trees, we extract a relationship matrix (right) that describes all the pairwise relationships and calculate F1 scores/accuracy by comparing the matrices.}
        \label{fig:evaluation}
    \end{center}
\end{figure}

While we call our system a ``transition parser'', we do not employ a stack and instead employ the graph-based parser-like formulation for the pointer identification.
We selected this strategy because of the recent success of graph-based parsers \cite{dozat-2017-deep,zhang-etal-2019-amr}.

\section{Implementation and Customization}\label{sec:implementation}

In this section, we briefly describe the implementation of our system that allows easy customization to different types of VSDs.
Our system employs modular and customizable design and is implemented in Python.
A user may implement a new feature extractor simply by writing a new feature extractor class where each feature is implemented as its class function (\cref{fig:implementation}).
For example, \texttt{@single\_input\_feature([1])} denotes that the subsequent function should be applied to the second block of each context (thus corresponding to feature V6).
Likewise, the features for pointer identification can be implemented by marking a function with \texttt{@pointer\_feature()}, which takes a candidate block $b_j$ (\texttt{tb1}), a target block $b_i$ (\texttt{tb2}), the block next to the target block $b_{i+1}$ (\texttt{tb3}) and $b_{head(j)}$ (\texttt{head\_tb}) as an input.

A feature extractor object is instantiated for each document where all feature functions are automatically aggregated to produce the feature vector.
A new feature extractor can inherit from an existing feature extractor (e.g., feature extractors for \contractpdfen and \contractpdfja both inherit from a base PDF feature extractor), which makes it easy to reuse implementations.

\begin{table*}[t!]
    \centering
    \begin{threeparttable}
        \centering
        \fontsize{8pt}{10pt}\selectfont
        \renewcommand{\arraystretch}{.6}
        \setlength\tabcolsep{1mm}
        \begin{tabularx}{\linewidth}{Xcccccp{3mm}cccp{3mm}cccp{3mm}ccc}\toprule
            & & & \multicolumn{3}{c}{\contractpdfen} & & \multicolumn{3}{c}{\lawpdfen}  & & \multicolumn{3}{c}{\contracttxten}  & & \multicolumn{3}{c}{\contractpdfja} \\\cmidrule(lr){4-6}\cmidrule(lr){8-10}\cmidrule(lr){12-14}\cmidrule(lr){16-18}
            Relationship & & & Visual & Number & Ours & & Visual & Number & Ours  & & Visual & Number & Ours  & & Visual & Number & Ours \\\midrule
            \arrayrulecolor{black!30}
            \multirow{6}{5mm}{Same paragraph} & \parbox[t]{1mm}{\multirow{3}{*}{\rotatebox[origin=c]{90}{Micro}}}
                & P & \textbf{0.982} & 0.484 & 0.944 & & \textbf{0.891} & 0.219 & 0.858 & & \textbf{0.993} & 0.540 & 0.983  &  & 0.446 & 0.402 & \textbf{0.973} \\
            &  & R & 0.683 & 0.947 & \textbf{0.951} & & 0.681 & \textbf{0.969} & 0.957 & & 0.708 & 0.917 & \textbf{0.978}  &  & 0.552 & \textbf{0.985} & 0.966 \\
            &  & F & 0.806 & 0.641 & \textbf{0.947} & & 0.772 & 0.357 & \textbf{0.905} & & 0.826 & 0.680 & \textbf{0.980}  &  & 0.494 & 0.571 & \textbf{0.969} \\\cmidrule{2-18}
            & \parbox[t]{1mm}{\multirow{3}{*}{\rotatebox[origin=c]{90}{Macro}}}
                & P & \textbf{0.980} & 0.644 & 0.955 & & 0.906 & 0.328 & \textbf{0.936} & & \textbf{0.990} & 0.595 & 0.969 &  & 0.481 & 0.478 & \textbf{0.971} \\
            &  & R & 0.670 & \textbf{0.966} & 0.951 & & 0.634 & \textbf{0.974} & 0.951 & & 0.746 & 0.934 & \textbf{0.976} &  & 0.527 & \textbf{0.985} & 0.956 \\
            &  & F &0.782 & 0.736 & \textbf{0.948} & & 0.731 & 0.452 & \textbf{0.933} & & 0.847 & 0.687 & \textbf{0.971} &  & 0.450 & 0.617 & \textbf{0.955} \\\midrule
            \multirow{6}{5mm}{Siblings} & \parbox[t]{1mm}{\multirow{3}{*}{\rotatebox[origin=c]{90}{Micro}}}
                & P & 0.332 & 0.677 & \textbf{0.841} & & 0.430 & 0.647 & \textbf{0.849} & & 0.397 & 0.780 & \textbf{0.784}  &  & 0.106 & 0.151 & \textbf{0.699} \\
            &  & R & 0.323 & \textbf{0.765} & 0.736 & & 0.283 & 0.504 & \textbf{0.712} & & 0.481 & \textbf{0.763} & 0.723 &  & 0.506 & 0.571 & \textbf{0.691} \\
            &  & F & 0.328 & 0.718 & \textbf{0.785} & & 0.341 & 0.567 & \textbf{0.774} & & 0.435 & \textbf{0.772} & 0.752  &  & 0.176 & 0.238 & \textbf{0.695} \\\cmidrule{2-18}
            & \parbox[t]{1mm}{\multirow{3}{*}{\rotatebox[origin=c]{90}{Macro}}}
                & P & 0.443 & 0.678 & \textbf{0.791} & & 0.598 & 0.493 & \textbf{0.793} & & 0.482 & 0.677 & \textbf{0.814} &  & 0.347 & 0.237 & \textbf{0.719} \\
            &  & R & 0.427 & 0.691 & \textbf{0.751} & & 0.417 & 0.379 & \textbf{0.696} & & 0.557 & 0.603 & \textbf{0.758} &  & 0.506 & 0.536 & \textbf{0.663} \\
            &  & F & 0.337 & 0.650 & \textbf{0.748} & & 0.410 & 0.385 & \textbf{0.724} & & 0.435 & 0.605 & \textbf{0.754} &  & 0.292 & 0.283 & \textbf{0.671} \\\midrule
            \multirow{6}{5mm}{Descendants} & \parbox[t]{1mm}{\multirow{3}{*}{\rotatebox[origin=c]{90}{Micro}}}
                & P & 0.381 & 0.184 & \textbf{0.502} & & \textbf{0.627} & 0.132 & 0.456 & & 0.239 & 0.190 & \textbf{0.541} &  & 0.536 & 0.125 & \textbf{0.577} \\
            &  & R & 0.123 & \textbf{0.879} & 0.807 & & 0.303 & \textbf{0.881} & 0.858 & & 0.048 & \textbf{0.888} & 0.771 &  & 0.340 & 0.580 & \textbf{0.826} \\
            &  & F & 0.186 & 0.304 & \textbf{0.619} & & 0.409 & 0.229 & \textbf{0.596} & & 0.080 & 0.313 & \textbf{0.635} &  & 0.416 & 0.205 & \textbf{0.679} \\\cmidrule{2-18}
            & \parbox[t]{1mm}{\multirow{3}{*}{\rotatebox[origin=c]{90}{Macro}}}
                & P & 0.295 & 0.242 & \textbf{0.655} & & 0.438 & 0.173 & \textbf{0.581} & & 0.193 & 0.269 & \textbf{0.639} &  & 0.462 & 0.122 & \textbf{0.737} \\
            &  & R & 0.194 & \textbf{0.848} & 0.798 & & 0.314 & 0.764 & \textbf{0.837} & & 0.072 & \textbf{0.859} & 0.735 &  & 0.358 & 0.519 & \textbf{0.834} \\
            &  & F & 0.203 & 0.340 & \textbf{0.641} & & 0.327 & 0.230 & \textbf{0.617} & & 0.096 & 0.367 & \textbf{0.625} &  & 0.372 & 0.195 & \textbf{0.739} \\\midrule
            \multirow{2}{20mm}{\qquad Accuracy} &   \multicolumn{2}{l}{Micro}
                      & 0.772 & 0.778 & \textbf{0.914} & & 0.827 & 0.685 & \textbf{0.908} & & 0.587 & 0.674 & \textbf{0.828} &  & 0.618 & 0.623 & \textbf{0.940} \\
            & \multicolumn{2}{l}{Macro}
                     & 0.686 & 0.679 & \textbf{0.889} & & 0.732 & 0.427 & \textbf{0.840} & & 0.571 & 0.580 & \textbf{0.841} &  & 0.623 & 0.492 & \textbf{0.899} \\\midrule
            \multirow{2}{20mm}{\qquad Average F1} &   \multicolumn{2}{l}{Micro}
                     & 0.440 & 0.555 & \textbf{0.784} & & 0.507 & 0.384 & \textbf{0.758} & & 0.447 & 0.588 & \textbf{0.789} &  & 0.362 & 0.338 & \textbf{0.781} \\
             & \multicolumn{2}{l}{Macro}
                     & 0.441 & 0.576 & \textbf{0.779} & & 0.489 & 0.356 & \textbf{0.758} & & 0.459 & 0.553 & \textbf{0.783} &  & 0.372 & 0.365 & \textbf{0.788} \\
            \arrayrulecolor{black}\bottomrule
        \end{tabularx}
        \makeatletter\def\TPT@hsize{}\makeatletter
        \begin{tablenotes}[para,flushleft]
            \raggedright
            \fontsize{8pt}{8pt}\selectfont
            ``Micro'': Micro-average, ``Macro'': Macro-average, ``P'': Precision, ``R'': Recall, ``F'': F1 score
        \end{tablenotes}
    \end{threeparttable}
    \caption{Results for evaluation on IE perspective}\label{tab:result-structure}
\end{table*}

\section{Experiments}\label{sec:experiments}

\subsection{Evaluation Metrics}

While we do report transition prediction accuracy, it is not a true task metric since it is rooted on our formulation of the task.
Looking back at our initial motivation in \cref{sec:introduction}, we introduce two sets of evaluation metrics.

The first set of metrics is rooted on IE perspective.
For IE, it is important to identify ancestor-descendant and sibling relationships because it allows, for example, identifying a subject (in an ancestral block) and its objects (a decendant block and its siblings).
Thus, we evaluate F1 scores for identifying pairs of blocks in
\begin{enumerate*}[label=(\arabic*)]
    \item same paragraph,
    \item sibling, and
    \item ancestor-descendant
\end{enumerate*} relationships, respectively (\cref{fig:evaluation}).
Note that we do not include cousin blocks in the sibling relationship, because it is not clear whether cousin blocks have any meaningful information in the context of IE.

We use the second set of metrics to evaluate a system's efficacy as a preprocessing tool for more general NLP pipelines.
We evaluate paragraph boundary identification metrics since paragraph boundaries can be used to determine appropriate chunks of text to be fed into the NLP pipelines.
We also report accuracy for removing debris with \texttt{omitted}.

We used five-folds cross validation for the evaluation.

\subsection{Baselines}

We compared our system against the following baselines:
\begin{description}
    \item[Numbering baseline]\cite{hatsutori_estimating_2017}
    This baseline detects numberings using a set of regular expressions and identifies dropping in hierarchy when the type of numberings has changed.
    Adopting \citet{hatsutori_estimating_2017} to our problem formulation, our implementation is the same as the feature ``numbering transition (T1).''
    \item[Visual baseline] This baseline relies purely on visual cues; i.e., indentation and line spacing.
    For each pair of consecutive blocks, this baseline outputs
    \begin{enumerate*}[label=(\arabic*)]
        \item \texttt{continuous} when indentation does not change and line spacing is normal (as in feature V8),
        \item \texttt{consecutive} when indentation does not change and line spacing is larger than normal,
        \item \texttt{down} when indentation gets larger, and
        \item \texttt{up} when indentation gets smaller
    \end{enumerate*}.
    On \texttt{up}, it points back at the closest block with the same indentation.
    \item[PDFMiner] We use this popular open-source project to detect paragraph boundaries as in \citet{bast_benchmark_2017}.
    PDFMiner relies purely on geometric heuristics to detect paragraph breaks.
\end{description}

\begin{table*}[t!]
    \centering
    \begin{threeparttable}
        \centering
        \fontsize{8pt}{10pt}\selectfont
        \renewcommand{\arraystretch}{.6}
        \setlength\tabcolsep{0.3mm}
        \begin{tabularx}{\linewidth}{Xccccccp{1mm}ccccp{1mm}cccp{1mm}cccc}\toprule
            & & & \multicolumn{4}{c}{\contractpdfen} & & \multicolumn{4}{c}{\lawpdfen}  & & \multicolumn{3}{c}{\contracttxten}  & & \multicolumn{4}{c}{\contractpdfja} \\\cmidrule(lr){4-7}\cmidrule(lr){9-12}\cmidrule(lr){14-17}\cmidrule(lr){18-21}
            Criteria & & & PDFMiner & Visual & Number & Ours & & PDFMiner & Visual & Number & Ours  & & Visual & Number & Ours  & & PDFMiner & Visual & Number & Ours \\\midrule
            \arrayrulecolor{black!30}
            \multirow{6}{5mm}{Paragraph boundary} & \parbox[t]{2.5mm}{\multirow{3}{*}{\rotatebox[origin=c]{90}{Micro}}}
                & P & 0.672 & 0.563 & 0.914 & \textbf{0.958} & & 0.546 & 0.536 & 0.911 & \textbf{0.948} & & 0.465 & 0.783 & \textbf{0.955} &  & 0.531 & 0.603 & 0.961 & \textbf{0.970} \\
            &  & R & 0.822 & \textbf{0.968} & 0.700 & 0.948 & & 0.858 & 0.916 & 0.637 & \textbf{0.948} & & \textbf{0.989} & 0.637 & 0.945 &  & 0.850 & 0.663 & 0.627 & \textbf{0.991} \\
            &  & F & 0.739 & 0.712 & 0.793 & \textbf{0.953} & & 0.667 & 0.676 & 0.750 & \textbf{0.948} & & 0.633 & 0.702 & \textbf{0.950} &  & 0.653 & 0.632 & 0.759 & \textbf{0.980} \\\cmidrule{2-21}
            & \parbox[t]{2.5mm}{\multirow{3}{*}{\rotatebox[origin=c]{90}{Macro}}}
                & P & 0.698 & 0.598 & 0.921 & \textbf{0.958} & & 0.632 & 0.565 & 0.866 & \textbf{0.946} & & 0.527 & 0.840 & \textbf{0.953} &  & 0.585 & 0.645 & 0.964 & \textbf{0.970} \\
            &  & R & 0.798 & \textbf{0.964} & 0.703 & 0.945 & & 0.874 & 0.930 & 0.522 & \textbf{0.943} & & \textbf{0.984} & 0.633 & 0.944 &  & 0.867 & 0.653 & 0.624 & \textbf{0.988} \\
            &  & F & 0.722 & 0.729 & 0.772 & \textbf{0.947} & & 0.703 & 0.692 & 0.620 & \textbf{0.940} & & 0.673 & 0.693 & \textbf{0.947} &  & 0.661 & 0.627 & 0.745 & \textbf{0.976} \\\midrule
            \multirow{6}{5mm}{Block elimination} & \parbox[t]{2.5mm}{\multirow{3}{*}{\rotatebox[origin=c]{90}{Micro}}}
                & P &  ---  &  ---  &  ---  & 0.969 & &  ---  &  ---  &  ---  & 0.979 & &  ---  &  ---  & 0.865 &  &  ---  &  ---  &  ---  & 1.000 \\
            &  & R &  ---  &  ---  &  ---  & 0.897 & &  ---  &  ---  &  ---  & 0.755 & &  ---  &  ---  & 0.914 &  &  ---  &  ---  &  ---  & 0.849 \\
            &  & F &  ---  &  ---  &  ---  & 0.932 & &  ---  &  ---  &  ---  & 0.852 & &  ---  &  ---  & 0.889 &  &  ---  &  ---  &  ---  & 0.919 \\\cmidrule{2-21}
            & \parbox[t]{2.5mm}{\multirow{3}{*}{\rotatebox[origin=c]{90}{Macro}}}
            & P &  ---  &  ---  &  ---  & 0.948 & &  ---  &  ---  &  ---  & 0.929 & &  ---  &  ---  & 0.815 &  &  ---  &  ---  &  ---  & 0.929 \\
            &  & R &  ---  &  ---  &  ---  & 0.906 & &  ---  &  ---  &  ---  & 0.858 & &  ---  &  ---  & 0.816 &  &  ---  &  ---  &  ---  & 0.866 \\
            &  & F &  ---  &  ---  &  ---  & 0.913 & &  ---  &  ---  &  ---  & 0.874 & &  ---  &  ---  & 0.800 &  &  ---  &  ---  &  ---  & 0.888 \\
            \arrayrulecolor{black}\bottomrule
        \end{tabularx}
        \makeatletter\def\TPT@hsize{}\makeatletter
        \begin{tablenotes}[para,flushleft]
            \raggedright
            \fontsize{8pt}{8pt}\selectfont
            ``Micro'': Micro-average, ``Macro'': Macro-average, ``P'': Precision, ``R'': Recall, ``F'': F1 score
        \end{tablenotes}
    \end{threeparttable}
    \caption{Results for evaluation on preprocessing perspective}\label{tab:result-preprocessing}
\end{table*}

\begin{table*}[tb]
    \centering
    \begin{threeparttable}
        \centering
        \fontsize{8pt}{10pt}\selectfont
        \setlength\tabcolsep{1mm}
        \renewcommand{\arraystretch}{.8}
        \begin{tabular}{ll@{\hskip 2 pt}ll@{\hskip 2 pt}ll@{\hskip 2 pt}ll@{\hskip 2 pt}ll@{\hskip 2 pt}ll@{\hskip 2 pt}ll@{\hskip 2 pt}ll@{\hskip 2 pt}l}\toprule
             & \multicolumn{4}{c}{\contractpdfen} & \multicolumn{4}{c}{\lawpdfen} & \multicolumn{4}{c}{\contracttxten} & \multicolumn{4}{c}{\contractpdfja}\\\cmidrule(lr){2-5}\cmidrule(lr){6-9}\cmidrule(lr){10-13}\cmidrule(lr){14-17}
            \# & \multicolumn{2}{c}{Forward} & \multicolumn{2}{c}{Backward} & \multicolumn{2}{c}{Forward} & \multicolumn{2}{c}{Backward} & \multicolumn{2}{c}{Forward} & \multicolumn{2}{c}{Backward} & \multicolumn{2}{c}{Forward} & \multicolumn{2}{c}{Backward}\\\midrule
            & All & (0.914) & All & (0.914) & All & (0.908) & All & (0.908) & All & (0.828) & All & (0.828) & All & (0.940) & All & (0.940) \\
            1 & T1, 2 & (0.763) & T1, 2 & (0.855)  &  T1, 2 & (0.685) & T1, 2 & (0.854) & V8, 2-3 & (0.333) & T2, 2 & (0.820) & T1, 2 & (0.596) & T1, 2 & (0.934) \\
            2 & V10, 2 & (0.796) & T10, 3 & (0.794)  &  V8, 2-3 & (0.883) & V10, 2 & (0.859) & T10, 2 & (0.465) & V9, 2 & (0.811) & V12, 2 & (0.686) & V9, 2 & (0.909) \\
            3 & T10, 3 & (0.818) & T10, 2 & (0.794)  &  V10, 2 & (0.893) & T2, 2 & (0.836) & T9, 2 & (0.716) & T3, 2 & (0.805) & V1, 2-3 & (0.821) & V8, 2-3 & (0.882) \\
            4 & T7, 3 & (0.853) & V1, 2-3 & (0.796)  &  V8, 1-2 & (0.885) & V8, 2-3 & (0.800) & T3, 2 & (0.727) & V5 & (0.785) & T2, 2 & (0.813) & S1\textsuperscript{\textdaggerdbl}, 2 & (0.865) \\
            5 & T10, 2 & (0.813) & S1\textsuperscript{\textdaggerdbl}, 2 & (0.808)  &  V5, 2-3 & (0.858) & V1, 2-3 & (0.747) & T6, 2 & (0.721) & T10, 2 & (0.781) & V8, 2-3 & (0.887) & T2, 2 & (0.856) \\
            6 & V1, 2-3 & (0.844) & T8, 2-3 & (0.801)  & T7, 3 & (0.881) & V4, 2-3 & (0.716) & T4, 2 & (0.723) & V8, 2-3 & (0.752) & V9, 3 & (0.906) & V4, 2 & (0.824) \\
            7 & V4, 2 & (0.868) & T2, 2 & (0.774)  &  V1, 2-3 & (0.898) & V9, 2 & (0.676) & T2, 2 & (0.721) & S1\textsuperscript{\textdagger}, 2 & (0.749) & T2, 1 & (0.913) & V1, 2-3 & (0.787) \\
            8 & T2, 2 & (0.886) & V4, 2-3 & (0.692)  &  V3, 2 & (0.904) & S1\textsuperscript{\textdagger}, 2 & (0.717) & V4, 2 & (0.722) & V9, 3 & (0.751) & V4, 2-3 & (0.926) & V9, 2 & (0.799) \\
            \bottomrule
        \end{tabular}
        \begin{tablenotes}[para,flushleft]
            \raggedright
            \fontsize{8pt}{8pt}\selectfont
            Numbers in parentheses show micro-average transition label prediction accuracy. The first line shows the results with all features. \textsuperscript{\textdagger}:~$\ell(i, i - 1) - \ell(i + 1, i - 1)$ variant. \textsuperscript{\textdaggerdbl}:~$\ell(i + 1, i) - \ell(i + 1, i - 1)$ variant.
        \end{tablenotes}
    \end{threeparttable}
    \caption{Eight most important features chosen by greedy forward selection and backward elimination.}\label{tab:feature-importance}
\end{table*}

\subsection{Implementation Details}

We used Random Forest \cite{breiman_2001} as the transition and pointer classifiers, which is suited for categorical features that occupy the majority of our features.
We did not tune hyperparameters of the Random Forest classifier and used default values of scikit-learn \cite{pedregosa_2011}.

For language model coherence feature S1, we used GPT-2 medium\footnote{\url{https://huggingface.co/gpt2}} for English documents and \texttt{japanese-gpt2-medium}\footnote{\url{https://huggingface.co/rinna/japanese-gpt2-medium}}) for Japanese documents.

\subsection{Results}

Structure and preprocessing evaluations are shown on \cref{tab:result-structure} and \cref{tab:result-preprocessing}, respectively.
Our system obtained micro-average structure prediction accuracy of 0.914 for \contractpdfen, 0.908 for \lawpdfen, 0.828 for \contracttxten and 0.940 for \contractpdfja, significantly outperforming the best baselines with 0.778, 0.827, 0.674 and 0.623, respectively.
Our system performed the best with respect to F1 scores for all but one structure relationships.

The difference was even more significant for paragraph boundary detection.
For \contractpdfen, our system obtained a micro-average paragraph boundary detection F1 score of 0.953 that is significantly better than PDFMiner with an F1 score of 0.739.
PDFMiner performed on par with our visual baseline and generally performed worse than our numbering baseline.
This shows the importance of incorporating textual information to preprocess VSDs.

Micro-average transition label prediction accuracies were 0.951 (\contractpdfen), 0.938 (\lawpdfen), 0.955 (\contracttxten) and 0.923 (\contractpdfja).

We investigated the importance of each feature with greedy forward selection and greedy backward elimination of the features (\cref{tab:feature-importance}).
We can observe that our system makes a balanced use of the visual and textual cues.
``Indentation (V1)'', ``Larger line spacing (V8)'' and ``numbering hierarchy (T1)'', which partially represent the baselines, were ranked high in many cases.
At the same time, other features such as ``all capital (T10)'' and ``punctuated (T2)'' were also contributing significantly to the accuracy, which made our system much superior to the baselines.

The feature importance revealed that the semantic cue (S1) was no more important than other cues.
We suspect that the feature (which compares whether adjacent or non-adjacent block is more likely given a context) had fallen back to mere language model with the context being ignored in some cases, possibly due to GPT-2 not being fine-tuned on the legal domain.

We also conducted a qualitative error analysis.
For \contractpdfen, we found that our system was performing poorly on documents where they had bold or underlined section titles, followed by paragraphs without any indentation (predicted \texttt{continuous} instead of \texttt{down}).
We believe incorporating typographic features would improve our system as implied by the success of the ``all capital (T10)'' feature.

For \contracttxten, we found that blocks that are all capitals or are all underbars were misclassified as \texttt{omitted}.
All capital words and underbars are frequently used to denote headers and footers, but they were used as section titles and input fields in these examples.
Unlike for \contractpdfen, we attribute this problem to lack of training data, as those should have been classified correctly with other features (such as T4 and T8) if the system had seen similar patterns in the training data.

Interestingly, we observed that the system tends to do better in documents that are hierarchically more complex.
This may be because hierarchically complex documents tend to incorporate more cues to support humans comprehend the documents.

\section{Related Works}\label{sec:related_works}

As discussed in \cref{sec:introduction}, previous works mainly focused on word segmentation and layout analysis, whereas fine-grained logical structure analysis of VSDs is less addressed.
Nevertheless, there exist some studies that focus on similar goals.

 \citet{abreu_findsefintoc-2019_2019} and \citet{ferres_pdfdigest_2018} have tried to deal with logical structure analysis by identifying specific structures in VSDs such as subheadings.
However, these studies are too coarse-grained and cannot handle paragraph-level logical structure, thus they are unable to satisfy the need we have discussed in \cref{sec:introduction}.
FinSBD-3 shared task \cite{au-2021-finsbd} is more fine-grained than those works and incorporates extraction of list items.
However, its main focus is not on analysis of logical structures; it has only four static levels for list hierarchies and does not consider hierarchies in non-list paragraphs.

\citet{hatsutori_estimating_2017} proposed a rule-based system that purely relies on numberings.
We compared our system against it in \cref{sec:experiments} and showed that our system, which also incorporates textual and semantic cues, is superior to their method.

\citet{sporleder_automatic_2004} proposed a paragraph boundary detection method for plain texts that purely relies on textual and semantic cues.
While their method is not intended for VSDs, some of their ideas could be incorporated to our work as additional features.
We leave use of more advanced semantic cues for a future work.

While the goal is different, our textual features have some similarity to those used in sentence boundary detection \cite{gillick-2009-sentence}.
Since our goal is to predict structures as well as boundaries, we employ richer textual and visual features that they do not utilize.

LayoutLM \cite{xu_layoutlm_2020,xu-etal-2021-layoutlmv2} incorporates multimodal self-supervised learning to utilize deep learning for form understanding.
While it may alleviate the need for a large training dataset, it is not trivial to adopt the same method for logical structure analysis as text blocks would not fit onto the LayoutLM's context.
Furthermore, it is easier to diagnose and to improve our system as it utilizes a combination of hand-crafted features, while deep learning systems tend to be completely black box.

\section{Conclusions}\label{sec:conclusion}

We proposed a transition parser-like formulation of the logical structure analysis of VSDs and developed a feature-based machine learning system that fuses visual, textual and semantic cues.
Our system significantly outperformed baselines and an existing open-source software on different types of VSDs.
The experiment revealed that incorporating both the visual and textual cues is crucial in successfully conducting logical structure analysis of VSDs.
As a future work, we will incorporate typographic and more advanced semantic cues.

\ifanonymous \else \section*{Acknowledgements}

We used computational resource of AI Bridging Cloud Infrastructure (ABCI) provided by the National Institute of Advanced Industrial Science and Technology (AIST) for the experiments.
 \fi

\bibliography{main}

\begin{thebibliography}{18}
\expandafter\ifx\csname natexlab\endcsname\relax\def\natexlab#1{#1}\fi

\bibitem[{Abreu et~al.(2019)Abreu, Cardoso, and
  Oliveira}]{abreu_findsefintoc-2019_2019}
Carla Abreu, Henrique Cardoso, and Eugénio Oliveira. 2019.
\newblock \href {https://www.aclweb.org/anthology/W19-6410}
  {{FinDSE}@{FinTOC}-2019 {Shared} {Task}}.
\newblock In \emph{Proceedings of the {Second} {Financial} {Narrative}
  {Processing} {Workshop}}.

\bibitem[{Au et~al.(2021)Au, Ait-Azzi, and Kang}]{au-2021-finsbd}
Willy Au, Abderrahim Ait-Azzi, and Juyeon Kang. 2021.
\newblock \href {https://doi.org/10.1145/3442442.3451378} {{FinSBD}-2021: {The}
  3rd {Shared} {Task} on {Structure} {Boundary} {Detection} in {Unstructured}
  {Text} in the {Financial} {Domain}}.
\newblock In \emph{Companion Proceedings of the Web Conference 2021}.

\bibitem[{Bast and Korzen(2017)}]{bast_benchmark_2017}
Hannah Bast and Claudius Korzen. 2017.
\newblock \href {https://doi.org/10.1109/JCDL.2017.7991564} {A {Benchmark} and
  {Evaluation} for {Text} {Extraction} from {PDF}}.
\newblock In \emph{2017 {ACM}/{IEEE} {Joint} {Conference} on {Digital}
  {Libraries}}.

\bibitem[{Breiman(2001)}]{breiman_2001}
Leo Breiman. 2001.
\newblock \href {https://doi.org/10.1023/A:1010933404324} {Random {Forests}}.
\newblock \emph{Machine Learning}, 45(1):5--32.

\bibitem[{Dozat and Manning(2017)}]{dozat-2017-deep}
Timothy Dozat and Christopher~D. Manning. 2017.
\newblock \href {https://openreview.net/forum?id=Hk95PK9le} {Deep {Biaffine}
  {Attention} for {Neural} {Dependency} {Parsing}}.
\newblock In \emph{5th International Conference on Learning Representations}.

\bibitem[{Ferrés et~al.(2018)Ferrés, Saggion, Ronzano, and
  Bravo}]{ferres_pdfdigest_2018}
Daniel Ferrés, Horacio Saggion, Francesco Ronzano, and Àlex Bravo. 2018.
\newblock \href {https://www.aclweb.org/anthology/L18-1298} {{PDFdigest}: an
  {Adaptable} {Layout}-{Aware} {PDF}-to-{XML} {Textual} {Content} {Extractor}
  for {Scientific} {Articles}}.
\newblock In \emph{Proceedings of the {Eleventh} {International} {Conference}
  on {Language} {Resources} and {Evaluation}}.

\bibitem[{Gillick(2009)}]{gillick-2009-sentence}
Dan Gillick. 2009.
\newblock \href {https://www.aclweb.org/anthology/N09-2061} {Sentence
  {Boundary} {Detection} and the {Problem} with the {U}.{S}.}
\newblock In \emph{Proceedings of Human Language Technologies: The 2009 Annual
  Conference of the North {A}merican Chapter of the Association for
  Computational Linguistics}.

\bibitem[{Hatsutori et~al.(2017)Hatsutori, Yoshikawa, and
  Imai}]{hatsutori_estimating_2017}
Yoichi Hatsutori, Katsumasa Yoshikawa, and Haruki Imai. 2017.
\newblock \href {https://doi.org/10.1007/978-3-319-61572-1_18} {Estimating
  {Legal} {Document} {Structure} by {Considering} {Style} {Information} and
  {Table} of {Contents}}.
\newblock In \emph{New {Frontiers} in {Artificial} {Intelligence}}, pages
  270--283. Springer International Publishing.

\bibitem[{Obermaier et~al.(2016)Obermaier, Obermayer, Wormer, and
  Jaschensky}]{obermaier_about_2016}
Frederik Obermaier, Bastian Obermayer, Vanessa Wormer, and Wolfgang Jaschensky.
  2016.
\newblock \href
  {https://panamapapers.sueddeutsche.de/articles/56febff0a1bb8d3c3495adf4/}
  {About the {Panama} {Papers}}.
\newblock \emph{Süddeutsche Zeitung}.

\bibitem[{Pedregosa et~al.(2011)Pedregosa, Varoquaux, Gramfort, Michel,
  Thirion, Grisel, Blondel, Prettenhofer, Weiss, Dubourg, Vanderplas, Passos,
  Cournapeau, Brucher, Perrot, and Duchesnay}]{pedregosa_2011}
F.~Pedregosa, G.~Varoquaux, A.~Gramfort, V.~Michel, B.~Thirion, O.~Grisel,
  M.~Blondel, P.~Prettenhofer, R.~Weiss, V.~Dubourg, J.~Vanderplas, A.~Passos,
  D.~Cournapeau, M.~Brucher, M.~Perrot, and E.~Duchesnay. 2011.
\newblock \href {https://www.jmlr.org/papers/v12/pedregosa11a.html}
  {Scikit-learn: {Machine} {Learning} in {P}ython}.
\newblock \emph{Journal of Machine Learning Research}, 12:2825--2830.

\bibitem[{Radford et~al.(2019)Radford, Wu, Child, Luan, Amodei, and
  Sutskever}]{radford2019language}
Alec Radford, Jeffrey Wu, Rewon Child, David Luan, Dario Amodei, and Ilya
  Sutskever. 2019.
\newblock \href
  {https://cdn.openai.com/better-language-models/language_models_are_unsupervised_multitask_learners.pdf}
  {Language {Models} are {Unsupervised} {Multitask} {Learners}}.
\newblock \emph{OpenAI blog}, 1(8):9.

\bibitem[{Soto and Yoo(2019)}]{soto_visual_2019}
Carlos Soto and Shinjae Yoo. 2019.
\newblock \href {https://doi.org/10.18653/v1/D19-1348} {Visual {Detection} with
  {Context} for {Document} {Layout} {Analysis}}.
\newblock In \emph{Proceedings of the 2019 {Conference} on {Empirical}
  {Methods} in {Natural} {Language} {Processing} and the 9th {International}
  {Joint} {Conference} on {Natural} {Language} {Processing}}.

\bibitem[{Sporleder and Lapata(2004)}]{sporleder_automatic_2004}
Caroline Sporleder and Mirella Lapata. 2004.
\newblock \href {https://www.aclweb.org/anthology/W04-3210} {Automatic
  {Paragraph} {Identification}: {A} {Study} across {Languages} and {Domains}}.
\newblock In \emph{Proceedings of the 2004 {Conference} on {Empirical}
  {Methods} in {Natural} {Language} {Processing}}.

\bibitem[{Stahl et~al.(2018)Stahl, Young, Herrmannova, Patton, and
  Wells}]{stahl_deeppdf_2018}
Christopher Stahl, Steven Young, Drahomira Herrmannova, Robert Patton, and Jack
  Wells. 2018.
\newblock \href
  {http://lrec-conf.org/workshops/lrec2018/W24/summaries/14_W24.html}
  {{DeepPDF}: {A} {Deep} {Learning} {Approach} to {Extracting} {Text} from
  {PDFs}}.
\newblock In \emph{Proceedings of the {Eleventh} {International} {Conference}
  on {Language} {Resources} and {Evaluation}}.

\bibitem[{{The U.S. Securities and Exchange Commission}(2018)}]{sec_edgar_2018}
{The U.S. Securities and Exchange Commission}. 2018.
\newblock \emph{EDGAR® Public Dissemination Service Technical Specification}.

\bibitem[{Xu et~al.(2021)Xu, Xu, Lv, Cui, Wei, Wang, Lu, Florencio, Zhang, Che,
  Zhang, and Zhou}]{xu-etal-2021-layoutlmv2}
Yang Xu, Yiheng Xu, Tengchao Lv, Lei Cui, Furu Wei, Guoxin Wang, Yijuan Lu,
  Dinei Florencio, Cha Zhang, Wanxiang Che, Min Zhang, and Lidong Zhou. 2021.
\newblock \href {https://doi.org/10.18653/v1/2021.acl-long.201}
  {{L}ayout{LM}v2: Multi-modal pre-training for visually-rich document
  understanding}.
\newblock In \emph{Proceedings of the 59th Annual Meeting of the Association
  for Computational Linguistics and the 11th International Joint Conference on
  Natural Language Processing}.

\bibitem[{Xu et~al.(2020)Xu, Li, Cui, Huang, Wei, and Zhou}]{xu_layoutlm_2020}
Yiheng Xu, Minghao Li, Lei Cui, Shaohan Huang, Furu Wei, and Ming Zhou. 2020.
\newblock \href {https://doi.org/10.1145/3394486.3403172} {Layoutlm:
  Pre-training of text and layout for document image understanding}.
\newblock In \emph{Proceedings of the 26th ACM SIGKDD International Conference
  on Knowledge Discovery \& Data Mining}.

\bibitem[{Zhang et~al.(2019)Zhang, Ma, Duh, and
  Van~Durme}]{zhang-etal-2019-amr}
Sheng Zhang, Xutai Ma, Kevin Duh, and Benjamin Van~Durme. 2019.
\newblock \href {https://doi.org/10.18653/v1/P19-1009} {{AMR} {Parsing} as
  {Sequence}-to-{Graph} {Transduction}}.
\newblock In \emph{Proceedings of the 57th Annual Meeting of the Association
  for Computational Linguistics}.

\end{thebibliography}
\bibliographystyle{acl_natbib}

\appendix
\clearpage

\section{Appendix}\label{sec:appendix}

\subsection{Details of Data Collection and Annotation}\label{sec:appendix-dataset}

In this section, we provide supplemental information regarding the data collection and the annotation discussed in \cref{sec:dataset}.

For PDFs, we queried Google search engines and downloaded the PDF files that the search engines returned.
We used the following queries and the domains:
\begin{description}
    \item[\contractpdfen] ``\,``non-disclosure'' agreement filetype:pdf'' on seven domains from countries where English is widely spoken (US ``.com'', UK ``.co.uk'', Australia ``.com.au'', New Zealand ``.co.nz'', Singapore ``.com.sg'', Canada ``.ca'', South Africa ``.co.za'').
    \item[\lawpdfen] ``site:*.gov ``order'' filetype:pdf'' on ``google.com''.
    \item[\contractpdfja] ````\begin{CJK}{UTF8}{min}秘密保持契約書\end{CJK}'' filetype:pdf'' on ``google.co.jp''.
\end{description}

For the collection of \contracttxten, we first download all the documents filed at EDGAR from 1996 to 2020 in a form of daily archives\footnote{\url{https://www.sec.gov/Archives/edgar/Oldloads/}}.
We uncompressed each archive and deserialized files using regular expressions by referencing to the EDGAR specifications\cite{sec_edgar_2018}, which gave us 12,851,835 filings each of which contains multiple documents.
We then extracted NDA candidates from the documents by a rule-based filtering.
Using meta-data obtained during the deserialization, we extracted documents whose file type starts with ``EX'' (denotes that it is an exhibit), its file extension is one of ``.pdf'', ``.PDF'', ``.txt'', ``.TXT'', ``.html'', ``.HTML'', ``.htm'' or ``HTM'', and its content is matched by a regular expression ``(?<![a-zA-Z\.,\-"()]\textvisiblespace*)([Nn]on[-\textvisiblespace][Dd]isclosure)|(NON[-\textvisiblespace]DISCLOSURE)''.

We then randomly selected documents that fulfill following criteria:\begin{itemize}
    \item it is an NDA or an executive order,
    \item it has embedded texts (for PDFs),
    \item it is a single column document, and
    \item a similar document is not yet in the dataset.
\end{itemize}
The last criterion mainly targets contracts from same organizations and executive orders from same authorities.
It ensures that we get a wide variety of documents in our dataset.

The datasets were annotated by one of the authors.
We did not employ majority vote to improve annotation consistency, because labels can be easily determined by a brief inspection of the document.

\end{document}